\documentclass[runningheads]{llncs}
\usepackage[T1]{fontenc}
\usepackage{graphicx}
\usepackage{booktabs}
\usepackage[misc]{ifsym}
\newcommand{\corr}{(\Letter)}
\usepackage{mwe}
\usepackage{amsmath}
\usepackage{amssymb}   
\usepackage{booktabs}
\usepackage{algorithm}

\usepackage{algpseudocode}
\usepackage{subcaption} 
\usepackage{tabularx} 

\usepackage{colortbl}
\usepackage[table]{xcolor}
\usepackage[hidelinks]{hyperref}
\usepackage{xspace}
\usepackage{multirow}
\newcommand{\ChordPrompt}{{\textsf{ChordPrompt}}\xspace}

\begin{document}

\title{ChordPrompt: Orchestrating Cross-Modal Prompt Synergy for Multi-Domain Incremental Learning in CLIP}

\titlerunning{ChordPrompt: Orchestrating Cross-Modal Prompt Synergy}

\author{Zhiyuan Wang\inst{1} \and
Bokui Chen\inst{1} \corr }

\authorrunning{Zhiyuan Wang and Bokui Chen}

\institute{Tsinghua Shenzhen International Graduate School, Tsinghua University, China \\ \email{\{wang-zy22, chenbk\}@tsinghua.edu.cn}
}

\maketitle              

\begin{abstract}
Continual learning (CL) empowers pre-trained vision-language models to adapt effectively to novel or previously underrepresented data distributions without comprehensive retraining, enhancing their adaptability and efficiency. While vision-language models like CLIP show great promise, they struggle to maintain performance across domains in incremental learning scenarios. Existing prompt learning methods face two main limitations: 1) they primarily focus on class-incremental learning scenarios, lacking specific strategies for multi-domain task incremental learning; 2) most current approaches employ single-modal prompts, neglecting the potential benefits of cross-modal information exchange. To address these challenges, we propose the \ChordPrompt framework, which facilitates a harmonious interplay between visual and textual prompts. \ChordPrompt introduces cross-modal prompts to leverage interactions between visual and textual information. Our approach also employs domain-adaptive text prompts to select appropriate prompts for continual adaptation across multiple domains. Comprehensive experiments on multi-domain incremental learning benchmarks demonstrate that \ChordPrompt outperforms state-of-the-art methods in zero-shot generalization and downstream task performance. Our implementation is currently available at \href{https://github.com/XiaoAI1989/ChordPrompt}{https://github.com/XiaoAI1989/ChordPrompt}.

\end{abstract}

\section{Introduction}

Continual Learning (CL) is a crucial paradigm in machine learning that aims to develop models capable of sequentially learning from various domains without the need for complete retraining from scratch. 

However, continual learning faces a critical challenge known as catastrophic forgetting \cite{french1999catastrophic}, which severely undermines the model's ability to master distinct tasks sequentially. Catastrophic forgetting occurs when neural networks lose their ability to perform previously learned tasks after training on new ones. This leads to a significant deterioration in performance on the initial tasks. This problem becomes particularly challenging when the model must adapt to new or under-represented data distributions, a common requirement in real-world deployments.

Vision-language models like CLIP \cite{radford2021learning} have shown remarkable performance on various multi-modal tasks, excelling in visual and linguistic knowledge. Consequently, they exhibit impressive zero-shot generalization performance on unseen datasets \cite{thengane2022clip}. Nevertheless, continually training Vision-Language (V-L) models like CLIP is critical. It helps keep the model up-to-date as new data emerges in real-world deployments. Unfortunately, during the continual fine-tuning of CLIP, its impressive zero-shot generalization performance substantially declines due to catastrophic forgetting. In addition, retraining large-scale vision-language models such as CLIP, pre-trained on 400M image-text pairs, for every new task would require computational resources often unavailable in real-world scenarios. 
Our approach provides a scalable solution by enabling continual learning without requiring access to the original training dataset or complete model retraining.

\begin{figure}[!htb]
    \centering
    \begin{subfigure}{0.8\textwidth}
        \includegraphics[width=\textwidth,keepaspectratio]{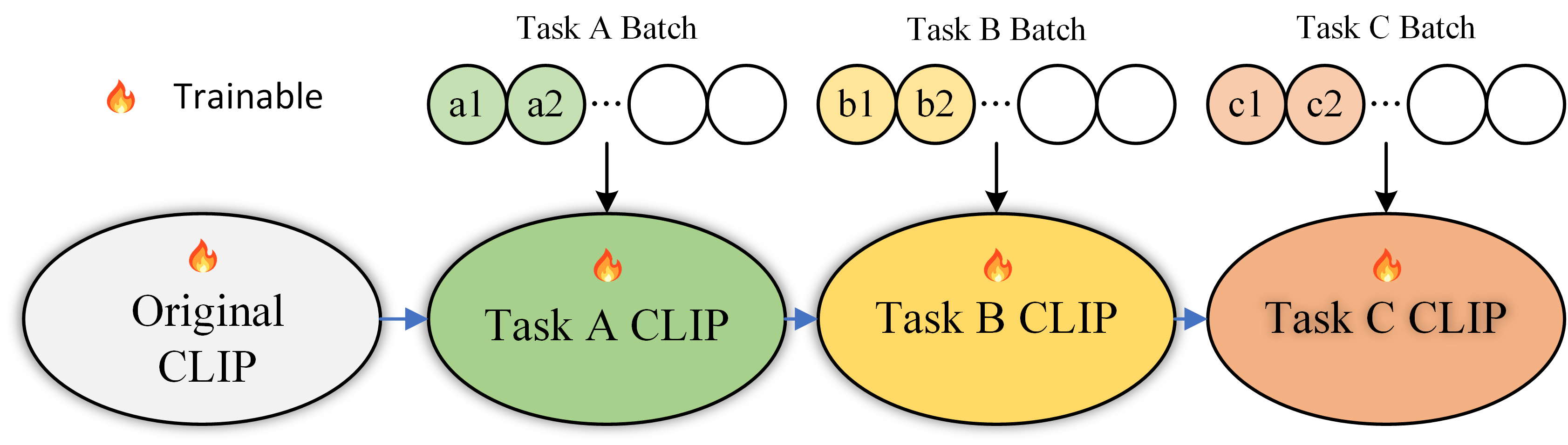}
        \caption{Traditional continual learning strategies require fine-tuning the entire model - a process that can be computationally expensive.}
        \label{fig:scene_a}
    \end{subfigure}
    \hfill
    \begin{subfigure}{0.8\textwidth}
        \includegraphics[width=\textwidth,keepaspectratio]{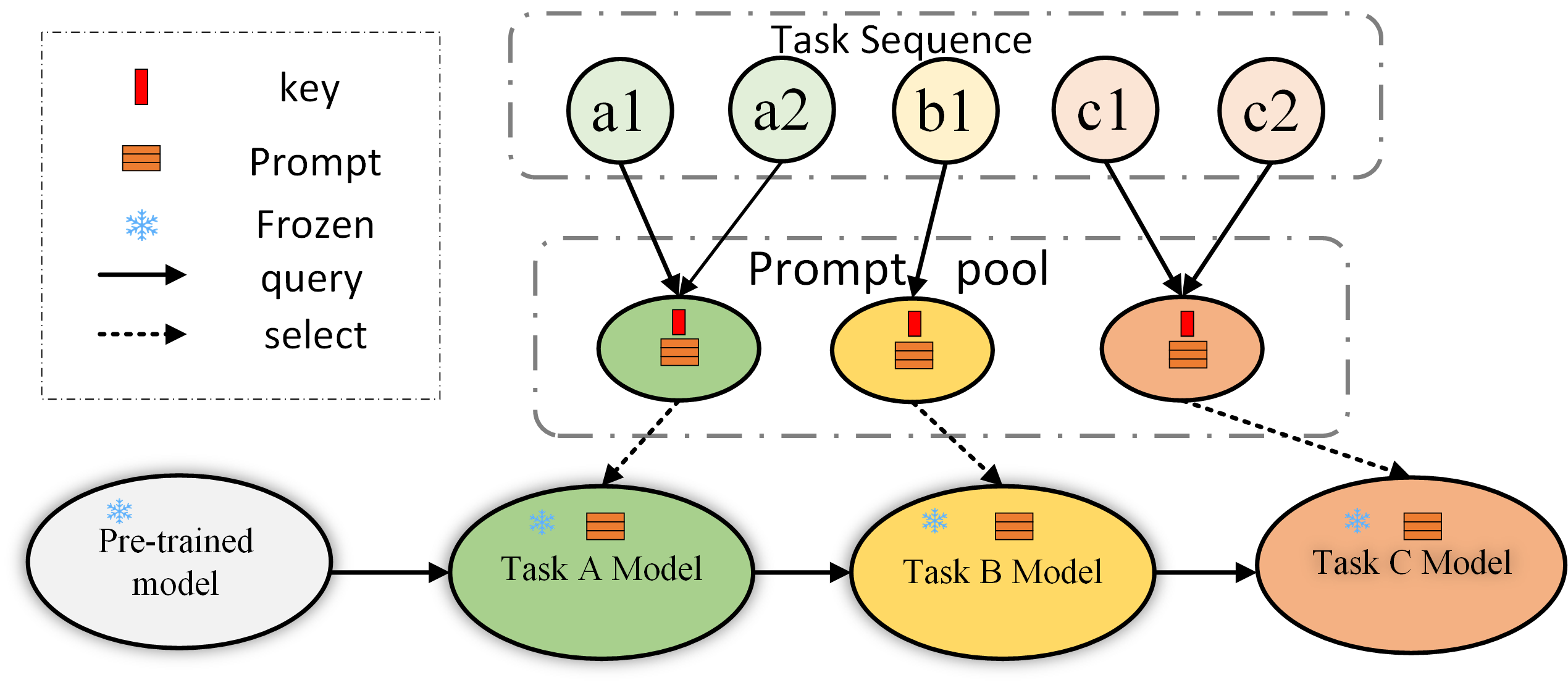}
        \caption{The single-modal prompt restricts the model to learning within the confines of a single modality, potentially missing valuable cross-modal insights.  }
        \label{fig:scene_b}
    \end{subfigure}
        \hfill
    \begin{subfigure}{0.8\textwidth}
        \includegraphics[width=\textwidth,keepaspectratio]{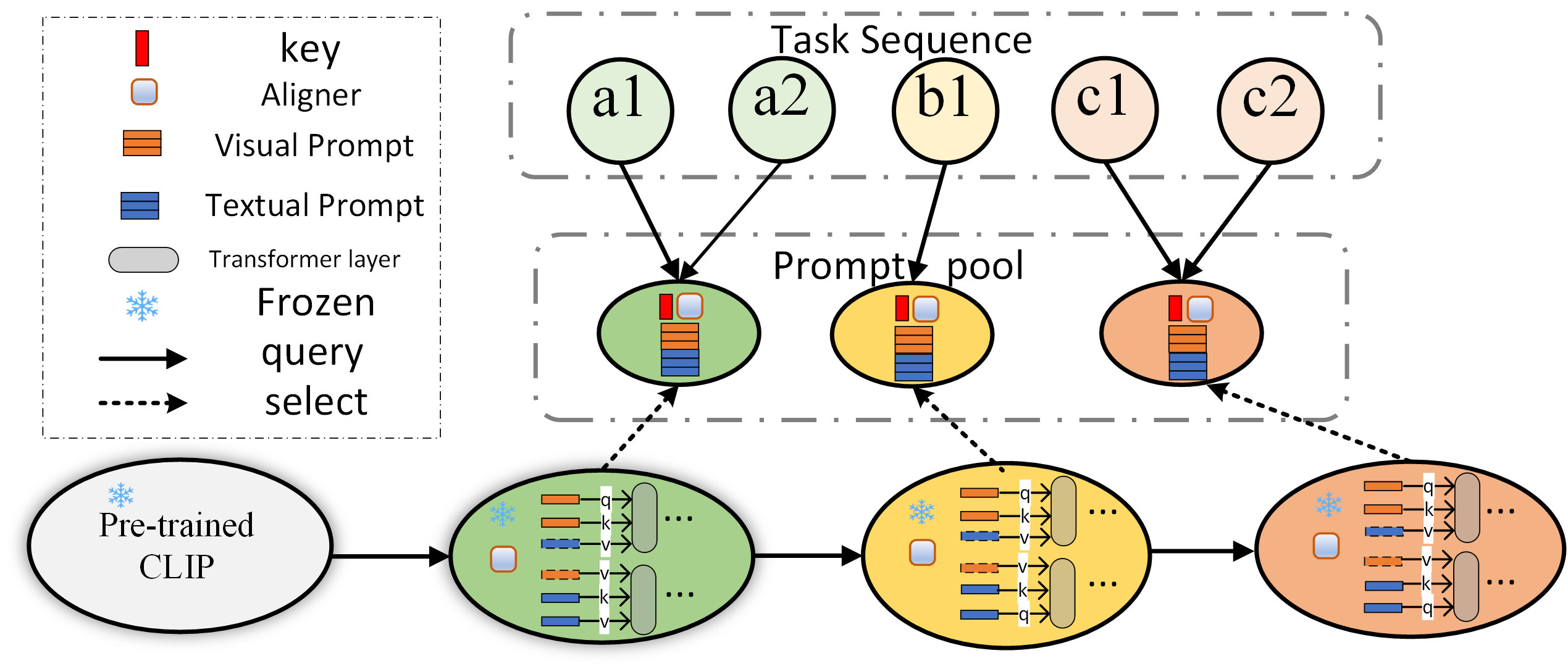}
        \caption{\ChordPrompt's cross-modal approach enables the model to benefit from the rich interplay between different modalities, leading to more robust and comprehensive learning. }
        \label{fig:scene_c}
    \end{subfigure}
    \caption{Comparison of traditional methods and \ChordPrompt.}
    \label{fig:scenes}
\end{figure}

Continual learning for vision-language models is an emerging field presenting many open challenges and opportunities. For the CLIP model, the use of replay methods is limited, as pre-training datasets are often private and inaccessible. Therefore, recent studies focus on fine-tuning the entire model \cite{Zheng_2023_ICCV,DBLP:conf/icml/NiWTZ023}. As shown in Figure \ref{fig:scene_a}, this method can inefficiently use both computational resources and the model's original capabilities.

To address these challenges, we introduce \ChordPrompt for Continual Learning framework. Our approach is motivated by the multi-modal hypothesis in cognitive psychology. This cognitive framework suggests that perception and learning are not isolated processes confined to individual sensory channels but somewhat interactive processes where information from one modality can modulate the processing of another \cite{shams2008benefits}. In traditional continual learning methods, as illustrated in Figure \ref{fig:scene_a}, the entire model typically requires fine-tuning. This process can be computationally expensive and may lead to inefficient use of the model's original capabilities. Contrastingly, as shown in Figure \ref{fig:scene_b}, single-modal prompts operate within the confines of a single modality, potentially missing valuable cross-modal insights. As shown in Figure \ref{fig:scene_c}, our proposed cross-modal approach \ChordPrompt leverages the synergistic relationship between visual and textual information, overcoming these limitations and enabling the model to learn and adapt more comprehensively across different scenarios. \ChordPrompt emulates the integrative nature of human cognition by fostering cross-modal interaction. This approach enables the model to develop more detailed and refined representations, significantly enhancing its performance in continual learning scenarios. Our work's main contributions are as follows:

\begin{itemize}
\item We introduce a cross-modal prompt strategy \ChordPrompt to facilitate continual learning in vision-language models.
\item We propose a domain-adaptive text prompt approach, which enables the model to adapt to specific characteristics of different domains, addressing the lack of strategies for multi-domain task incremental learning.
\item We design an Aligner module and cross-modal prompts in visual and textual encoders, addressing the limitation of single-modal prompts and leveraging the benefits of cross-modal information exchange.
\item We demonstrate \ChordPrompt's effectiveness across various tasks and datasets, particularly highlighting its performance in multi-domain task incremental learning scenarios with vision-language models. 
\end{itemize}

\section{Related Work}
\label{sec:related}

\textbf{Vision-Language Models.} Beyond CLIP, other V-L models like ALIGN, ViLBERT have shown effectiveness in various tasks. These models employ different architectures to process and integrate visual and textual information, ranging from unified representations to two-stream approaches with cross-modality layers ~\cite{jia2021scaling,lu2019vilbert}. The recent advancements in V-L models have played a crucial role in empowering a wide range of tasks that involve image and text processing \cite{liu2024visual,li2023blip,alayrac2022flamingo}. However, these V-L models often face challenges when deployed in continual learning scenarios. Many studies have proposed continual learning methods to aid the adaptation of the model to new tasks. Our proposed framework, \ChordPrompt, is tailored to continual learning (CL) setups for vision-language models, particularly those with dual-encoder architectures like CLIP. We chose CLIP as our base model due to its robust performance in aligning image and text embeddings and its widespread adoption and proven effectiveness in a variety of multimodal tasks like VQA.

\textbf{Prompt Learning Methods.} 
Prompt learning originating from NLP has become crucial in improving vision-language model performance through carefully selecting input prompts. For example, approaches like soft prompting \cite{lester2021power} and prefix tuning \cite{li2021prefix} help streamline large language models' training by introducing learnable tokens that steer the model's outputs, making it more convenient to tailor the model to new downstream tasks. In the context of vision-language models, prompt learning has been utilized to guide the models towards better understanding and alignment of visual and language modalities. For instance, CoOp and Co-CoOp ~\cite{zhou2022learning,zhou2022conditional} train CLIP for few-shot transfer by prompt vectors at the language branch. However, these single modal prompts can limit the model's ability to adapt to new tasks dynamically. Recent work \cite{xing2023dual,khattak2023maple,zhu2023visual} has begun to explore the potential of cross-modal prompt learning to fully leverage the power of the cross-modal nature of vision-language models. However, current cross-modal prompting methods lack ways to select appropriate prompts for continual adaptation. They typically lack the adaptability and flexibility needed to handle the dynamic shifts in the environment during continual learning.

\textbf{Continual Learning Methods.}  Multiple approaches are put forward to address the issue of catastrophic forgetting in continual learning. Replay-based methods~\cite{rebuffi2017icarl,shin2017continual,ding2022don,DBLP:conf/icml/JeeveswaranBZA23} store and replay previous data to maintain knowledge. Notably, in the case of the CLIP model, replay methods to re-access original private pre-training datasets face limitations, as these datasets are often inaccessible. Regularization-based methods~\cite{li2017learning,dhar2019learning,douillard2020podnet} mitigate catastrophic forgetting by aligning the current output with previous ones. Architecture-based methods \cite{rusu2016progressive,yoon2018lifelong} manipulate the model's architecture, such as dynamically expanding capacity or allocating model parts to each task. Prompt-based methods ~\cite{wang2022learning,wang2022s,douillard2022dytox,jung2023generating} have emerged to mitigate catastrophic forgetting. However, these prompt-based approaches ignore preserving the zero-shot learning capability, a crucial strength of vision-language models. Furthermore, existing approaches lack mechanisms for domain-adaptive selection, making them unsuitable for multi-domain task-incremental learning scenarios. Unlike previous methods primarily focusing on single-modal prompts or architectural modifications, our \ChordPrompt framework introduces a novel cross-modal prompting strategy to facilitate continual learning.

\section{Methodology}

\begin{figure*}[htb]
    \centering
    \includegraphics[width=\textwidth,keepaspectratio]{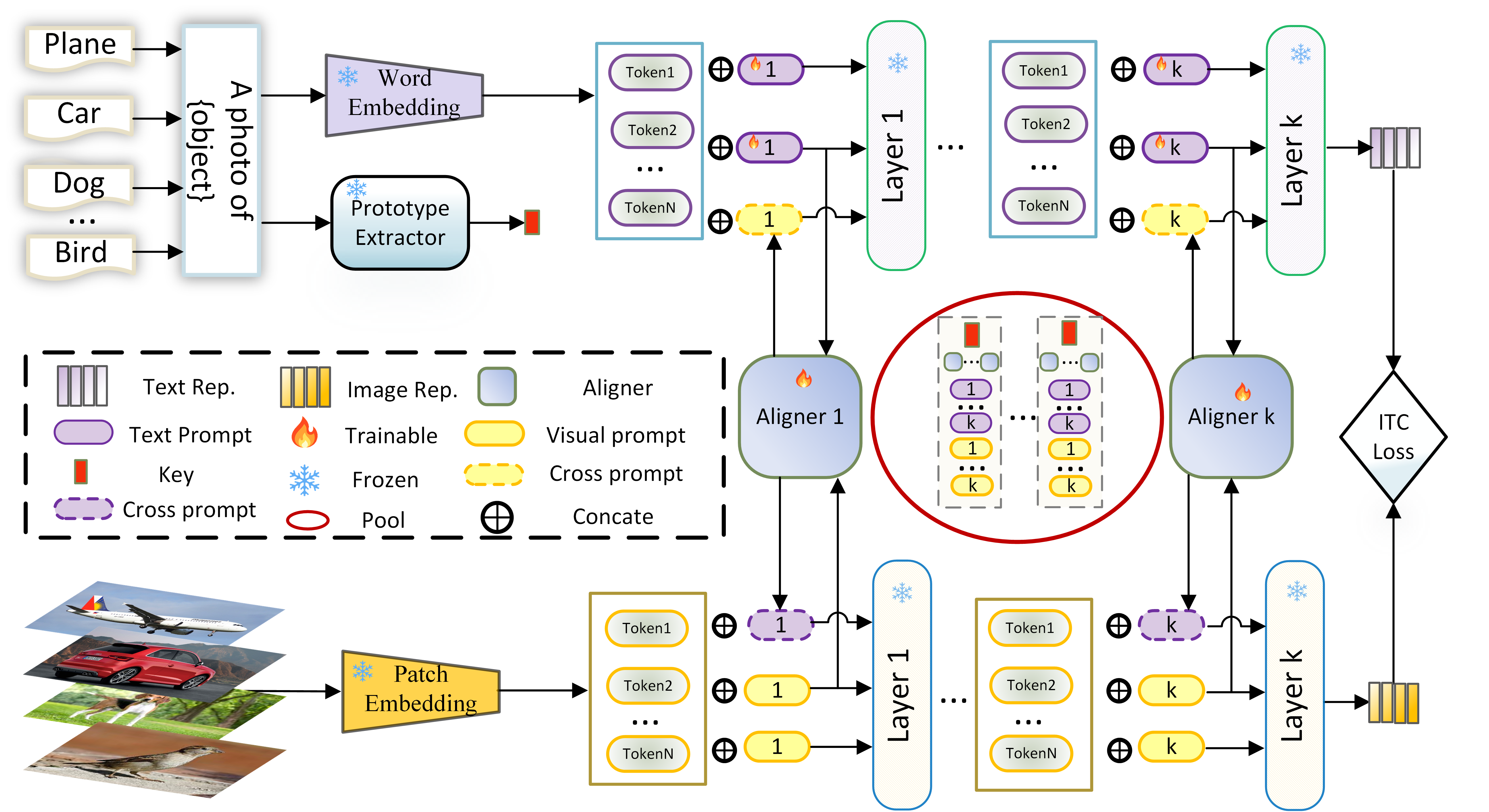}
    \caption{Our proposed \ChordPrompt approach. The detailed training process follows (a) Start by feeding all labels' text from the current dataset into the $\mathrm{Prototype Extractor}$. It transforms the text labels into one-dimensional vectors as keys. (b) Store these keys in a pool, which will later be used for querying during the inference phase. (c) Add trainable tokens as prompts at each layer of the text encoder, which are then stored in the memory pool. (d) Introduce the $\mathrm{Aligner}$ module, which projects text encoder prompts to the vision encoder and vision encoder prompts to the text encoder. The projected cross-modal prompts are added to each layer's value (V) component in both encoders. }
    \label{fig:alg}
\end{figure*}

In the field of continual learning (CL) for vision-language models, the goal is to develop models capable of sequentially mastering a series of distinct datasets, denoted as $\{\mathcal{D}_1, \mathcal{D}_2, \ldots, \mathcal{D}_{N}\}$. The $\mathcal{D}_{i}$ represents the dataset for the $i^{th}$ task, containing $N_i$ labeled samples. For multi-domain task-incremental learning, the objective is for the model to classify images from all domains it has encountered without knowing the specific task ID during inference. The optimization problem can be formulated as:
\small
\begin{equation}
\min_{\theta} \mathbb{E}_{(x, y)\sim \bigcup_{i=1}^{N}\mathcal{D}_i}\left[-\log \frac{\exp(s_{\theta}(x, \hat{y})/\tau)}{\sum_{y_c \in \mathbb{C}} \exp(s_{\theta}(x, y_c)/\tau)}\right]
\end{equation}
\normalsize
Here, $\mathcal{D}_i = {(x_j^i, \hat{y}_c^i)}_{j=1}^{N_i}$ represents the set of data for domain $\mathcal{D}_i$, which contains $N_i$ input-label pairs. $x_j^i$ is the $j^{th}$ input sample of domain $\mathcal{D}_i$, and $\hat{y}_c^i$ is the corresponding class label. The set of classes is denoted as $\mathbb{C} = \bigcup_{c=1}^{N_c} y_c$. \( N_c \) represents the total number of classes, and \( y_c \) denotes a specific class label within the set of classes. $\theta$ denotes the parameters of the model $F_\theta$. The loss function $\mathcal{L}$ measures the predictive discrepancy, traditionally the negative log-likelihood of the correct class in the context of contrastive learning.

In our CL formulation, we continually fine-tune the CLIP model on a sequence of datasets $\{\mathcal{D}_1, \mathcal{D}_2, ..., \mathcal{D}_{N}\}$.

The similarity score $s_{j,c}^i$, between visual $X_j^i$ and textual $Y_c^i$ representations is defined as $s_{j,c}^i = \text{sim}(X_j^i, Y_c^i)$. The prediction probability is calculated as follows:

\begin{equation}
p(\hat{y}|x_j^i) = \frac{\exp(s_{j,\hat{y}}^i/\tau)}{\sum_{c=1}^{N_c} \exp(s_{j,c}^i/\tau)}
\end{equation}
where $\tau$ represents the temperature coefficient.

\begin{algorithm}[htb]
\caption{Training Process for ChordPrompt}
\label{alg:training_ChordPrompt}
\begin{algorithmic}[1]
\Require
    Sequence of datasets $\{D_1, D_2, ..., D_N\}$;
    Pre-trained image encoder and text encoder;
    Trainable $\mathrm{Aligner}$ with $\theta_A$;
    Prototype extractor $\mathrm{ProtoExtrac}$;
    Text Prompt parameters $\theta_T$;
    Vision Prompt parameters $\theta_V$;
    Temperature coefficient $\tau$;
    
\Ensure
    Updated CLIP parameters

\For{$i = 1$ to $N$}
    \State Initialize trainable prompts $T_i$ and $V_i$
    \State Load dataset $D_i$ with samples $\{(x_j, y_j)\}_{j=1}^{N_i}$
    \State  $K_i \gets \mathrm{ProtoExtrac}(D_i)$ 
    \State Add key $K_i$ to prompt pool $\mathcal{P}$
    \For{$t = 1$ to $T_{\mathrm{iter}}$}
        \For{each $(x_j, y_j)$ in $D_i$}
            \State $\hat{T}_{i}, \hat{V}_{i} \gets \mathrm{Aligenr}(V_{i}; T_i)$
            \State  $Y_{i} \gets \mathrm{TextEncoder}(y_j; T_i, \hat{T}_{i})$
            \State  $X_{i,j} \gets \mathrm{ImageEncoder}(x_j; V_{i},\hat{V}_{i})$
            \State  $s_{i,j} \gets \mathrm{sim}(X_{i,j}, Y_{i})$
            \State Compute loss $\mathcal{L} \gets \mathcal{L}_{\mathrm{CE}}(s_{i,j}, y_j; \tau)$
            \State  $\theta_A, \theta_V, \theta_T \gets \mathrm{GradientDescent}(\mathcal{L}, \eta)$
        \EndFor
    \EndFor
    \State Update prompt pool $\mathcal{P}$ with new prompts $T_i$ and $V_i$ and key $K_i$
\EndFor

\State \Return $\mathcal{P}$
\end{algorithmic}
\end{algorithm}

\subsection{Domain-Adaptive Cross-modal Text Prompt}

Domain-adaptive text prompts are designed to dynamically adjust to the unique characteristics of each domain during training and inference. By associating each task with a prototype feature, these prompts ensure that the model can retrieve domain-specific knowledge, enabling task-specific adaptation and robust transfer across diverse domains.

CLIP's text encoder converts text into feature representations by tokenizing input text and projecting tokens into word embeddings $\mathbf{E}_0^i \in \mathbb{R}^{N_t \times d_t}$.

We introduce a novel Domain-Adaptive Text Prompt method to advance the conventional continual learning process. This methodology augments CLIP's text encoder with learnable tokens $T_l^{i}$ in $l^{th}$ layer of text encoder, each $T_l^{i} \in \mathbb{R}^{d_t}$, with lengths aligned to the visual prompts for cross-modality correspondence. Consequently, the input embeddings are a concatenation of the learnable prompts: $[\mathbf{E}_0^i, T_1^i]$, where $\mathbf{E}_0^i$ denotes the static input tokens. 

Specifically, we add new learnable tokens to each transformer layer $\mathrm{TextLayer}_l(\cdot)$ in the text encoder, up to $L$ layers in depth. The output embeddings $\mathbf{E}_{l-1}^i$ are sequentially fed into the $l^{th}$ transformer layer $\mathrm{TextLayer}_l$, for $l = 1, 2, \cdots, L$.

We introduce an $\mathrm{Aligner}$ module that projects visual prompts into the text space to enhance cross-modal interaction. The projected visual prompt $\hat{T}_l^i$ is computed as:

\begin{equation}
\hat{T}_l^i = A_{\mathrm{V2T}} V_l^i
\label{equ:2}
\end{equation}
where $A_{\mathrm{V2T}} \in \mathbb{R}^{d_t \times d_v}$ is a learnable matrix that aligns the visual prompt space to the text prompt space.

We incorporate the projected visual prompt into the value (V) component of the self-attention mechanism in the text encoder. This design choice is particularly effective because it does not interfere with the core attention computation. Specifically, Query (Q) and Key (K) are used to compute attention weights, so altering them could disrupt the attention alignment, whereas modifying V does not affect this crucial calculation. This approach preserves original attention patterns in Q and K while enhancing cross-modal information integration, allowing efficient information flow from the visual to the textual domain without disrupting the established attention mechanisms.




The text encoder is then modified to incorporate both the original text prompt and the projected visual prompt:

\small
\begin{equation}
\mathbf{E}_{l}^{i} = \mathrm{TextLayer}_l([\mathbf{E}_{l-1}^i, T_{l}^i, \hat{T}_l^i]), \quad l = 1, 2, \ldots, L.
\label{equ:3}
\end{equation}
\normalsize



Upon reaching the $L^{th}$ layer, the final textual representation $Y^i$ is obtained as:

\begin{equation}
Y^i = \mathrm{TextProj}\left(\mathbf{E}_{L}^{i}\right).
\label{equ:4}
\end{equation}
When the learnable tokens are only introduced at the initial textual embedding,  our approach is similar in structure to the CoOp method \cite{zhou2022learning}, which uses learnable class templates to replace manually designed templates. Our approach distinguishes itself through domain-adaptive prompts, which empower the model to flexibly adapt its responses to the unique characteristics of each domain, ultimately enabling more efficient continual learning.

\subsubsection{Prototype Extractor}
\begin{figure}[htb]
    \centering
    \includegraphics[width=0.8\textwidth,keepaspectratio]{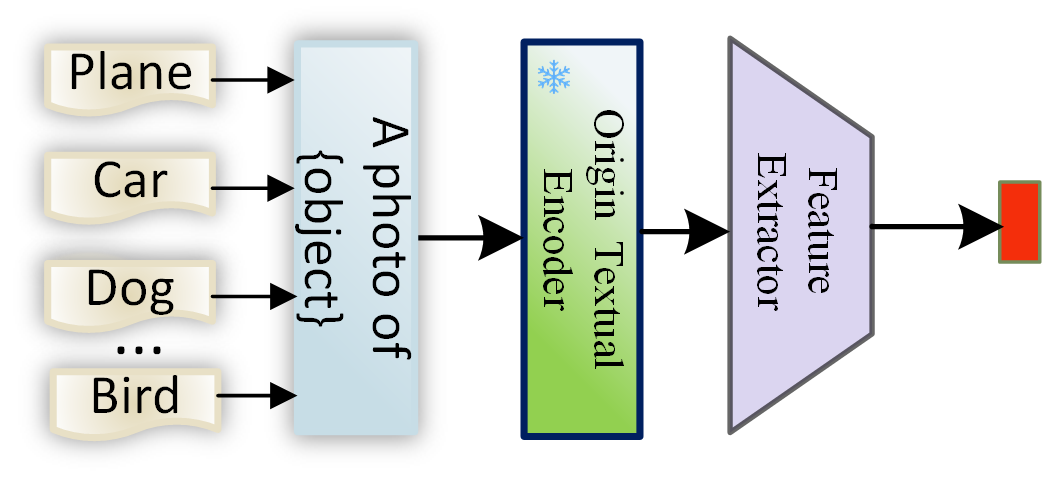}
    \caption{The architecture of our Prototype Extractor.}
    \label{fig:prototype}
\end{figure}

To capture the essence of each task's textual characteristics and to enable efficient retrieval of relevant prompts during inference, we introduce the Prototype Extractor. As can be seen from Figure \ref{fig:prototype}, we utilize the original CLIP model's text encoder $\mathrm{TextEnc}$ to acquire text feature representation $\hat{Y}_c^{i}$. Prototypes serve as compact representations of each task's textual characteristics. They enable efficient retrieval of relevant prompts during inference, ensuring the model can quickly adapt to the current task without losing previously acquired knowledge. The process is as follows:

\begin{equation}
\hat{Y}_{c}^{i} = \mathrm{TextEnc}(y_c^i)
\end{equation}
Herein, $\hat{Y}_{c}^{i}$ represents the feature embedding for the $c^{th}$ class of task $i$.

The prototype feature for task $i$, denoted as $P^i$, is computed as follows:

\begin{equation}
P^{i} = \frac{\sum_{c=1}^{N_c} Y_{c}^i}{\left| \sum_{c=1}^{N_c} Y_{c}^i \right|}
\end{equation}
In this equation, $P^{i}$ is the normalized prototype feature for the $i^{th}$ task. This is derived by calculating the sum of the feature representations $Y_{c}^i$ for all classes $c$ in task $i$ and then normalizing this sum to unit length. The normalization ensures that the prototype feature vector norm is one. These prototype features and their corresponding domain-specific prompts are stored in a prompt pool. The memory pool is designed to store only compact prototype features and corresponding prompts rather than raw data or task-specific checkpoints. This significantly reduces storage requirements and ensures scalability, even as the number of tasks increases. During inference, the model selects the prompt with the most similar key based on the maximum cosine similarity between the input and stored prototype features.

\subsection{Cross-modal Visual Prompt}

Cross-modal prompts with mixed visual and textual information are essential in vision-language models. By incorporating visual and projected text prompts, \ChordPrompt enables richer cross-modal interactions, leading to more robust and adaptable representations.

Image patches $U_b^j$ of $x_j^i$, where $N_b$ is the total number of patches, are initially embedded as $\mathbf{H}_{0,j}^i$:

\begin{equation}
\mathbf{H}_{0,j}^i = \mathrm{Embedding}(U_1^j,...,U_{N_b}^j)
\label{equ:8}
\end{equation}


To facilitate cross-modal learning, we introduce an $\mathrm{Aligner}$ module that projects text prompts into the vision space. The $\mathrm{Aligner}$ module, a key innovation in \ChordPrompt, facilitates bidirectional information flow between visual and textual modalities, enabling more robust and comprehensive representations. The projected text prompt $\hat{V}_l^i$ is computed as:

\begin{equation}
\hat{V}_l^i = A_{\mathrm{T2V}} T_l^i
\end{equation}
where $A_{\mathrm{T2V}} \in \mathbb{R}^{d_v \times d_t}$ is a learnable matrix that aligns the text prompt space to the visual prompt space.

The vision encoder is then modified to incorporate both the original visual prompt and the projected text prompt:

\small
\begin{equation}
\mathbf{H}_{l,j}^i = \mathrm{VisLayer}_l([\mathbf{H}_{l-1,j}^i, V_{l}^i, \hat{V}_l^i]), \quad l = 1, 2, \ldots, L. 
\end{equation}
\normalsize
Similar to the text encoder, we incorporate the projected text prompt into the value (V) component of the self-attention mechanism in the vision encoder. 

Upon reaching the $L^{th}$ layer, the prompts, and embeddings are combined to get the final visual representation $X^i_j$:

\begin{equation}
X^i_j = \mathrm{VisProj}\left(\mathbf{H}_{L,j}^{i}\right).
\label{equ:12}
\end{equation}
where $\mathrm{VisProj}$ is the visual projection layer, transforms the output of the visual encoder into a visual representation that can be directly compared with text embeddings in a shared embedding space.

In the end, We store the learned prompts and their corresponding keys for each task in a memory pool. We use cosine similarity during inference to retrieve the most suitable prompt from the pool based on the current task's input.

\begin{algorithm}[htb]
\caption{Inference Process for \ChordPrompt}
\label{alg:inference_ChordPrompt}
\begin{algorithmic}[1]
\Require
    The dataset class categories $\mathbb{C}$=$\{y_1,...,y_{N_c}\}$;
    Image $x$ for inference;
    Prototype extractor $\mathrm{ProtoExtrac}$;
    Discrimination threshold $\gamma$;
    Prompt pool $\mathcal{P}$ with keys $K_i$ and domain-specific prompts;
    Original model $\mathrm{CLIP}$;
    $\mathrm{Aligner}$ matrices $A_{\mathrm{V2T}}$, $A_{\mathrm{T2V}}$;

\Ensure
    Predicted class label $\hat{y}$ for image $x$;

\State $P_x \gets \mathrm{ProtoExtrac}(y_1,...,y_{N_c})$
\Statex \# Query the prompt pool and select the prompts
\State $\mathcal{S}_{\text{max}} \gets -\infty; i_{\text{max}} \gets \text{null}$
\ForAll{$K_i \in \mathcal{P}$}
    \State $\mathcal{S} \gets \frac{P_x \cdot K_i}{\lVert P_x \rVert \lVert K_i \rVert}$
    \If{$\mathcal{S} > \mathcal{S}_{\text{max}}$} 
        \State $\mathcal{S}_{\text{max}} \gets \mathcal{S}$ ; $i_{\text{max}} \gets i$ \hfill \# $i$ is the index of the key
    \EndIf
\EndFor
\Statex \# Inference using original CLIP
\If{$\mathcal{S}_{\text{max}} < \gamma$}
  \State  \Return $\hat{y}=\mathrm{CLIP}(x,y_1, ..., y_{N_c})$
\Else  
    \State $V^{i_{\text{max}}}, T^{i_{\text{max}}}, A_{\mathrm{V2T}}, A_{\mathrm{T2V}} \gets \mathcal{P}[i_{\text{max}}]$  \hfill \# Retrieve prompts
    \State $\hat{T}^{i_{\text{max}}} \gets A_{\mathrm{V2T}}V^{i_{\text{max}}}$  \hfill \# Project visual prompt to text space
    \State $\hat{V}^{i_{\text{max}}} \gets A_{\mathrm{T2V}}T^{i_{\text{max}}}$  \hfill \# Project text prompt to visual space
\EndIf 

\State $Y^{i_{\text{max}}} \gets \text{TextEncoder}(\mathbb{C}, T^{i_{\text{max}}}, \hat{T}^{i_{\text{max}}})$ \hfill \# Generate text representation
\State $X^{i_{\text{max}}} \gets \text{ImageEncoder}(x, V^{i_{\text{max}}}, \hat{V}^{i_{\text{max}}})$ \hfill \# Generate image representation

\State $\hat{y} \gets \arg\max(\mathrm{softmax}(\mathrm{sim}(Y^{i_{\text{max}}}, X^{i_{\text{max}}}) / \tau))$

\State \Return $\hat{y}$
\end{algorithmic}
\end{algorithm}

\subsection{Algorithm Architecture}

The detailed training process of our algorithm is illustrated in \ref{alg:training_ChordPrompt}.  
The detailed inference process is presented in Algorithm \ref{alg:inference_ChordPrompt}. Notably, our approach's image encoder of CLIP differs from that in conventional continual learning methods, as the latter do not utilize prompts in the vision branch. The more distinguishable image embeddings of \ChordPrompt emphasize that adding complementary visual prompts and domain-level text prompts leads to better adaptation of CLIP in continual learning scenarios.

We can define our learning objective to optimize the parameters for independent visual prompts $V$, separate textual prompts $T$, and the parameters of the $\mathrm{Aligner}$ component $\theta_A$. The learning objective can be formulated using the CE loss function. This optimization can be concisely represented as:
\begin{equation}
\min_{V, T, \theta_P} \mathcal{L}_{\text{CE}}\left(\mathcal{F}(x; V, T, \theta_A), y\right)
\end{equation}
where $\mathcal{L}_{\text{CE}}$ is the CE loss \cite{oord2018representation} that measures the predictive discrepancy in a contrastive learning setup, $F$ is the function representing the vision-language model parameterized by the prompts, and the $\mathrm{Aligner}$, $x$ is the input data, and $y$ is the label or target data used for contrastive prediction.

By minimizing this objective, \ChordPrompt optimizes the visual and textual prompts and the $\mathrm{Aligner}$ parameters. This approach enables the model to adapt to new tasks while preserving knowledge from previous tasks, effectively addressing catastrophic forgetting in continual learning settings. The CE loss ensures the model learns discriminative features across different modalities and tasks.

\begin{table*}[!htb]

  \setlength{\tabcolsep}{1.2pt}
  \begin{tabularx}{\textwidth}{lc|ccccccccccc|c}
    \hline
    Method & \rotatebox[origin=c]{90}{Param.} & \rotatebox[origin=c]{90}{Aircraft} & \rotatebox[origin=c]{90}{Caltech101} & \rotatebox[origin=c]{90}{CIFAR100} & \rotatebox[origin=c]{90}{DTD} & \rotatebox[origin=c]{90}{EuroSAT} & \rotatebox[origin=c]{90}{Flowers} & \rotatebox[origin=c]{90}{Food} & \rotatebox[origin=c]{90}{MNIST} & \rotatebox[origin=c]{90}{OxfordPet} & \rotatebox[origin=c]{90}{Cars} & \rotatebox[origin=c]{90}{SUN397} & Average  \\
    \hline
    CLIP ViT-b/16 \\
    Zero-shot & - & 24.3 & 88.4 & 68.2 & 44.6 & 54.9 & 71.0 & 88.5 & 59.4 & 89.0 & 64.7 & 65.2 & 65.3 \\
    full fine-tuning & - & 62.0 & 96.2 & 89.6 & 79.5 & 98.9 & 97.5 & 92.7 & 99.6 & 94.7 & 89.6 & 81.8 & 89.2\\
    \hline
    \textbf{Transfer} \\
    Continual-FT  & 211M & - & 67.1 & 46.0 & 32.1 & 35.6 & 35.0 & 57.7 & 44.1 & 60.8 & 20.5 & 46.6 & 44.6 \\
    LwF \cite{li2017learning} & 211M  & - & 74.5 & 56.9 & 39.1 & 51.1 & 52.6 & 72.8 & 60.6 & 75.1 & 30.3 & 55.9 &58.9\\
    iCaRL \cite{rebuffi2017icarl} & 211M  & - & 56.6 & 44.6 & 32.7 & 39.3 & 46.6 & 68.0 & 46.0 & 77.4 & 31.9 & 60.5 & 50.4\\
    LwF-VR \cite{ding2022don} & 211M  & - & 77.1 & 61.0 & 40.5 & 45.3 & 54.4 & 74.6 & 47.9 & 76.7 & 36.3 & 58.6 & 57.2\\
    WiSE-FT \cite{wortsman2022robust} & 211M  & - & 73.5 & 55.6 & 35.6 & 41.5 & 47.0 & 68.3 & 53.9 & 69.3 & 26.8 & 51.9 & 52.3\\
    Dist. only & 211M  & - & 80.1 & 62.2 & 40.2 & 39.9 & 58.1 & 80.8 & 53.4 & 74.6 & 38.1 & 61.9 & 58.9\\
    ZSCL \cite{Zheng_2023_ICCV} & 211M  & - & 86.0 & 67.4 & 45.4 & 50.4 & 69.1 & 87.6 & \textbf{61.8} & 86.8 & 60.1 & \textbf{66.8} & 68.1\\
    DDAS\cite{yu2024boosting}& 59.8M & - & 87.9 & 68.2 & 44.4 & 49.9 & 70.7 & \textbf{88.7} & 59.7 & \textbf{89.1} & 64.5 & 65.5  & 68.9\\

    \rowcolor{gray!25}
    \textbf{\ChordPrompt} & 9.5M  & - & \textbf{88.9} & \textbf{68.6} & \textbf{45.6} & \textbf{54.0} & \textbf{71.1} & 88.5 & 59.9 & 89.0 & \textbf{64.8} & 64.8  & \textbf{69.5} \\
    \hline

    \textbf{Avg.} \\
    Continual-FT & 211M  & 25.5 & 81.5 & 59.1 & 53.2 & 64.7 & 51.8 & 63.2 & 64.3 & 69.7 & 31.8 & 49.7 & 55.9 \\
    LwF \cite{li2017learning} & 211M  & 36.3 & 86.9 & 72.0 & 59.0 & 73.7 & 60.0 & 73.6 & 74.8 & 80.0 & 37.3 & 58.1 & 64.7\\
    iCaRL \cite{rebuffi2017icarl} & 211M  & 35.5 & 89.2 & 72.2 & 60.6 & 68.8 & 70.0 & 78.2 & 62.3 & 81.8 & 41.2 & 62.5 & 65.7\\
    LwF-VR \cite{ding2022don} & 211M  & 29.6 & 87.7 & 74.4 & 59.5 & 72.4 & 63.6 & 77.0 & 66.7 & 81.2 & 43.7 & 60.7 &65.1\\
    WiSE-FT \cite{wortsman2022robust} & 211M  & 26.7 & 86.5 & 64.3 & 57.1 & 65.7 & 58.7 & 71.1 & 70.5 & 75.8 & 36.9 & 54.6 & 60.7\\
    Dist. only & 211M  & 48.1 & 90.6 & 79.8 & 63.2 & 75.6 & 72.5 & 84.7 & 70.2 & 79.8 & 46.9 & 63.7 &70.5\\
    ZSCL \cite{Zheng_2023_ICCV} & 211M  & 45.1 & 92.0 & 80.1 & 64.3 & 79.5 & 81.6 & 89.6 & \textbf{75.2} & 88.9 & 64.7 & \textbf{68.0} & 75.4\\
    DDAS\cite{yu2024boosting} & 59.8M & 50.2 & 91.9 & \textbf{83.1} & 69.4 & 78.9 & 84.0 & 89.1 & 73.7 & 89.3 & 67.7 & 66.9 & 76.7\\
        \rowcolor{gray!25}
    \textbf{\ChordPrompt} & 9.5M  & \textbf{54.5} & \textbf{96.9} & 82.0 & \textbf{70.3} & \textbf{82.1} & \textbf{84.5} & \textbf{90.1} & 74.1 & \textbf{90.5} & \textbf{68.1} & 66.1 & \textbf{78.1}\\
    \hline
    
    \textbf{Last} \\
    Continual-FT & 211M  & 31.0 & 89.3 & 65.8 & 67.3 & 88.9 & 71.1 & 85.6 & 99.6 & 92.9 & 77.3 & 81.1 & 77.3 \\
    LwF \cite{li2017learning} & 211M  & 26.3 & 87.5 & 71.9 & 66.6 & 79.9 & 66.9 & 83.8 & 99.6 & 92.1 & 66.1 & 80.4 & 74.6 \\
    iCaRL \cite{rebuffi2017icarl} & 211M  & 35.8 & 93.0 & 77.0 & 70.2 & 83.3 & 88.5 & 90.4 & 86.7 & 93.2 & 81.2 & \textbf{81.9} & 80.1 \\
    LwF-VR \cite{ding2022don} & 211M  & 20.5 & 89.8 & 72.3 & 67.6 & 85.5 & 73.8 & 85.7 & 99.6 & 93.1 & 73.3 & 80.9 & 76.6 \\
    WiSE-FT \cite{wortsman2022robust} & 211M  & 27.2 & 90.8 & 68.0 & 68.9 & 86.9 & 74.0 & 87.6 & 99.6 & 92.6 & 77.8 & 81.3 & 77.7 \\
    Dist. only & 211M  & 43.3 & 91.9 & 81.3 & 72.4 & 95.1 & 90.5 & 90.4 & \textbf{99.7} & 92.5 & 85.1 & 81.8 & 84.0 \\
    ZSCL \cite{Zheng_2023_ICCV} & 211M   & 40.6 & 92.2 & 81.3 & 70.5 & 94.8 & 90.5 & 91.9 & 98.7 & 93.9 & \textbf{85.3} & 80.2  & 83.6 \\
    DDAS\cite{yu2024boosting} & 59.8M &  49.8 & 92.2 & \textbf{86.1} & 78.1 & 95.7 & 94.3 & 89.5 & 98.1 & 89.9 & 81.6 & 80.0  &85.0\\ 

    \rowcolor{gray!25}
    \textbf{\ChordPrompt} & 9.5M  & \textbf{54.5} & \textbf{97.1} & 85.0 & \textbf{79.5} & \textbf{98.2} & \textbf{95.8} & \textbf{92.0} & 99.1 & \textbf{94.4} & 83.0 & 79.0 & \textbf{87.0}\\  
    \hline

   \end{tabularx}
    \caption{Transfer, Average, and Last accuracy (\%) of various continue learning approaches on MTIL benchmark.}
    \label{tab:mtil}
\end{table*}

\section{Experiments}
\subsection{Datasets and Models}

\textbf{MTIL Benchmark.} Given that different classes from a single dataset usually have a common image source and a similar style \cite{hendrycks2021many,van2019three}, we suggest a cross-domain version of task incremental learning called Multi-domain Task Incremental Learning (MTIL). This benchmark presents a significant challenge for continual learning methods, as it requires the model to adapt to multiple domains while preserving its performance on previous tasks. In this framework, various tasks are gathered from distinct domains, each necessitating unique domain knowledge for humans to obtain high precision. Our MTIL benchmark comprises 11 tasks, including several tasks depicted in Table \ref{tab:mtil}. 

The MTIL benchmark presents a significant challenge, with 1,201 classes.
We employ a fixed sequence for evaluation. The datasets in Order I are organized in alphabetical sequence. Conversely, the Table \ref{table:table5} uses a random order
(Order-II): StanfordCars, Food, MNIST, OxfordPet, Flowers, SUN397, Aircraft, Caltech101, DTD, EuroSAT, CIFAR100. Experiments are done in Order I by default. Order II simulates real-world scenarios with unpredictable task arrivals, testing the robustness of domain-adaptive prompts to task order variations.

\textbf{Models.} We implement the CLIP model with a ViT-B/16 image encoder \cite{dosovitskiy2020image} and optimize it using the AdamW optimizer \cite{loshchilov2018decoupled}. We used a learning rate of 2e-3 for each task and a batch size of 64. We allocated 2000 iterations per task for multi-domain task incremental learning and followed the evaluation framework outlined in \cite{Zheng_2023_ICCV}. 

\begin{table*}[!htb]

  \setlength{\tabcolsep}{1.2pt}
  \begin{tabularx}{\textwidth}{lc|ccccccccccc|c}
    \hline
    Method & \rotatebox[origin=c]{90}{Param.} & \rotatebox[origin=c]{90}{Aircraft} & \rotatebox[origin=c]{90}{Caltech101} & \rotatebox[origin=c]{90}{CIFAR100} & \rotatebox[origin=c]{90}{DTD} & \rotatebox[origin=c]{90}{EuroSAT} & \rotatebox[origin=c]{90}{Flowers} & \rotatebox[origin=c]{90}{Food} & \rotatebox[origin=c]{90}{MNIST} & \rotatebox[origin=c]{90}{OxfordPet} & \rotatebox[origin=c]{90}{Cars} & \rotatebox[origin=c]{90}{SUN397} & Average \\
    \hline

\hline
    \textbf{Transfer} \\

Continual-FT & 211M & - & 72.8 & 53.0 & 36.4 & 35.4 & 43.3 & 68.4 & 47.4 & 72.6 & 30.0 & 52.7 & 51.2 \\
LwF \cite{li2017learning} & 211M  & - & 72.1 & 49.2 & 35.9 & 44.5 & 41.1 & 66.6 & 50.5 & 69.0 & 19.0 & 51.7 & 50.0 \\
LwF-VR\cite{ding2022don} & 211M  & - & 82.2 & 62.5 & 40.1 & 40.1 & 56.3 & 80.0 & 60.9 & 77.6 & 40.5 & 60.8 & 60.1 \\
WiSE-FT\cite{wortsman2022robust} & 211M  & - & 77.6 & 60.0 & 41.3 & 39.4 & 53.0 & 76.6 & 58.1 & 75.5 & 37.3 & 58.2 & 57.7 \\
ZSCL\cite{Zheng_2023_ICCV} & 211M  & - & 84.0 & 68.1 & 44.8 & 46.8 & 63.6 & 84.9 & 61.4 & 81.4 & 55.5 & 62.2 & 65.3 \\
DDAS\cite{yu2024boosting} & 59.8M  & - & 87.9 & 68.2 & 44.1 & 48.1 & 64.7 & \textbf{88.8} & \textbf{69.0} & \textbf{89.1} & 64.5 & \textbf{65.1} & 68.9 \\
    \rowcolor{gray!25}
\ChordPrompt & 9.5M  & - & \textbf{88.5} & \textbf{68.6} & \textbf{45.6} & \textbf{54.0} & \textbf{71.1} & 88.5 & 59.9 & 89.0 & \textbf{64.8} & 64.9 & \textbf{69.5} \\
\hline
\textbf{Avg.} \\

Continual-FT & 211M  & 28.1 & 86.4 & 59.1 & 52.8 & 55.8 & 62.0 & 70.2 & 64.7 & 75.5 & 35.0 & 54.0 & 58.5 \\
LwF \cite{li2017learning} & 211M  & 23.5 & 77.4 & 43.5 & 41.7 & 43.5 & 52.2 & 54.6 & 63.4 & 68.0 & 21.3 & 52.6 & 49.2 \\
LwF-VR \cite{ding2022don} & 211M  & 24.9 & 89.1 & 64.2 & 53.4 & 54.3 & 70.8 & 79.2 & 66.5 & 79.2 & 44.1 & 61.6 & 62.5 \\
WiSE-FT \cite{wortsman2022robust} & 211M  & 32.0 & 87.7 & 61.0 & 55.8 & 68.1 & 69.3 & 76.8 & 71.5 & 77.6 & 42.0 & 59.3 & 63.7 \\
ZSCL\cite{Zheng_2023_ICCV} & 211M  & 28.2 & 88.6 & 66.5 & 53.5 & 56.3 & 73.4 & 83.1 & 56.4 & 82.4 & 57.5 & 62.9 & 64.4 \\
DDAS\cite{yu2024boosting} & 59.8M  & 30.0 & 89.6 & 73.9 & 58.7 & 69.3 & 79.3 & \textbf{88.1} & \textbf{76.5} & 89.1 & 65.3 & \textbf{65.8} & 71.4 \\
    \rowcolor{gray!25}
\ChordPrompt & 9.5M & \textbf{39.3} & \textbf{93.8} & \textbf{74.5} & \textbf{59.2} & \textbf{75.0} & \textbf{83.0} & 87.8 & 71.4 & \textbf{89.8} & \textbf{66.7} & 65.6 & \textbf{73.3} \\
\hline
\textbf{Last} \\
Continual-FT & 211M  & 27.8 & 86.9 & 60.1 & 58.4 & 56.6 & 75.7 & 73.8 & 93.1 & 82.5 & 57.0 & 66.8 & 67.1 \\
LwF \cite{li2017learning} & 211M  & 22.1 & 58.2 & 17.9 & 32.1 & 28.1 & 66.7 & 46.0 & 84.3 & 64.1 & 31.5 & 60.1 & 46.5 \\
LwF-VR \cite{ding2022don} & 211M  & 22.9 & 89.8 & 59.3 & 57.1 & 57.6 & 79.2 & 78.3 & 77.7 & 83.6 & 60.1 & 69.8 & 66.9 \\
WiSE-FT\cite{wortsman2022robust} & 211M  & 30.8 & 88.9 & 59.6 & 60.3 & 80.9 & 81.7 & 77.1 & \textbf{94.9} & 83.2 & 62.8 & 70.0 & 71.9 \\
ZSCL\cite{Zheng_2023_ICCV} & 211M  & 26.8 & 88.5 & 63.7 & 55.7 & 60.2 & 82.1 & 82.6 & 58.6 & 85.9 & 66.7 & 70.4 & 67.4 \\
DDAS\cite{yu2024boosting} & 59.8M   & 30.0 & 89.6 & 73.9 & 58.7 & 69.3 & 79.3 & \textbf{88.1} & 76.5 & 89.1 & 65.3 & 65.8 & 71.4 \\
    \rowcolor{gray!25}
\ChordPrompt & 9.5M & \textbf{39.3} & \textbf{94.1} & \textbf{75.8} & \textbf{64.3} & \textbf{87.0} & \textbf{92.9} & 87.0 & 91.6 & \textbf{91.8} & \textbf{75.3} & \textbf{72.8} & \textbf{79.3} \\
\hline
   \end{tabularx}
\caption{Comparison with state-of-the-art methods on few-shot MTIL benchmark in terms of "Transfer", "Average", and "Last" scores (\%). Ours converges in 500 iterations on few-shot. We label the best and second methods with bold and underline styles. The top block indicates the upper-bound solutions to adapt the CLIP on each task.}
\label{tab:few_shot_mtil_comparison}
\end{table*}

\textbf{Metrics.} The measures for MTIL are displayed in Table \ref{tab:mtil}, where the rows indicate training steps, and each column represents the performance for a specific dataset. For traditional continual learning, only the scores below the diagonal of the accuracy matrix carry significance since they do not allow for zero-shot predictions on unknown tasks. However, the zero-shot transfer capability of vision-language models allows them to generate predictions across all datasets. The \textbf{"Avg"} metric represents the mean accuracy across all datasets evaluated at every training step, providing an overall measure of the model's performance throughout the continual learning process. The \textbf{"Last"} metric represents the performance of every task after the continual learning process, indicating the model's flexibility in adapting to downstream tasks. The \textbf{"Transfer"} metric is computed as the average task performance in the upper-right triangle of the accuracy matrix, assessing the model's ability to maintain its zero-shot transfer capability before learning task $i$, disregarding tasks learned after task $i$. A model that excels in both the "Last" and "Transfer" metrics exemplifies the ideal continual learner, adapting to new tasks while retaining past knowledge and generalization abilities.


\subsection{Ablation Study}

\begin{figure}[htb]
\centering
\begin{subfigure}{0.47\textwidth}
\centering
\includegraphics[width=\linewidth]{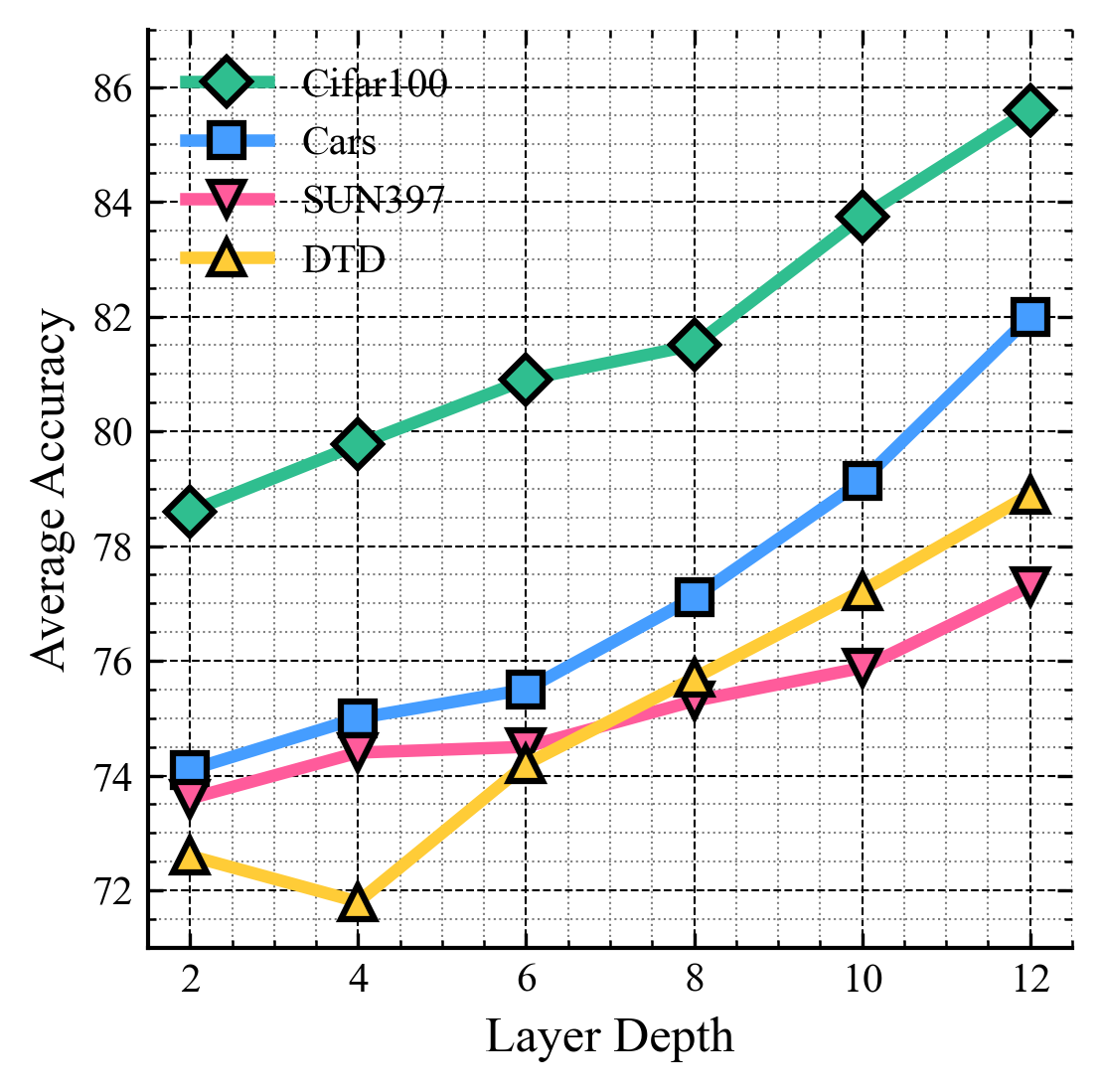}
\caption{The relationship between accuracy and the layer depth. (Prompt length=2. For better readability, we only display four tasks with close accuracy.)}
\label{fig:img1}
\end{subfigure}
\hfill
\begin{subfigure}{0.47\textwidth}
\centering
\includegraphics[width=\linewidth]{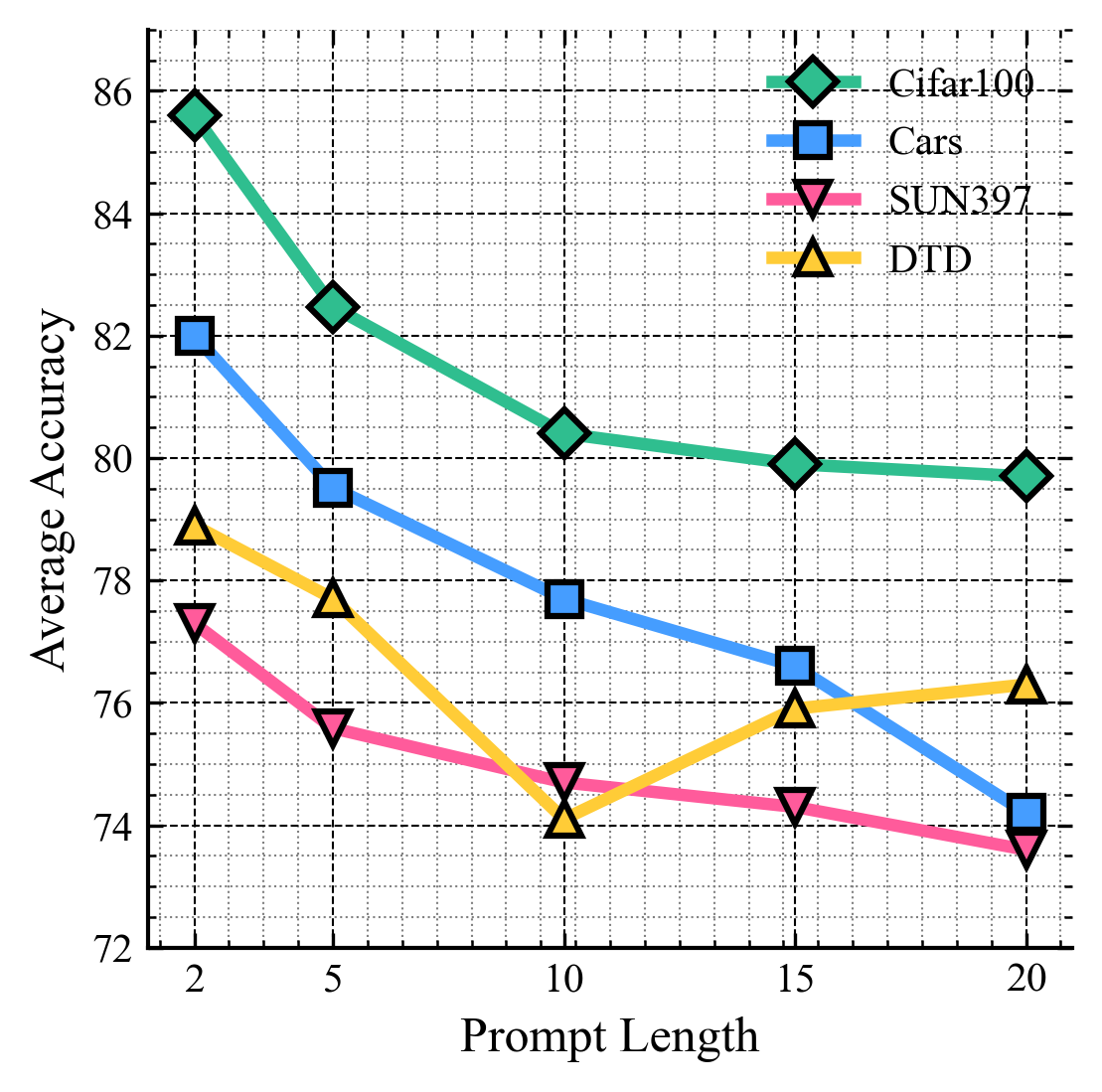}
\caption{The relationship between accuracy and the prompt length. (layer depth=12. For better readability, we only display four tasks with close accuracy.)}
\label{fig:img2}
\end{subfigure}
\caption{Comparisons of Performance}
\label{fig:performance_comparison}
\end{figure}

\textbf{Layer Depth.} As shown in Figure \ref{fig:img1}, we examine the effect of different layer depths on the \ChordPrompt methodology. As the model's feature space is more mature and stable, inserting prompts into deeper layers of the frozen model leads to a less significant impact. Therefore, we add the prompts in a front-to-back manner. 
In this context, \ChordPrompt generally attains optimal performance with a layer depth of 12 across most datasets. A shallower depth of 10 leads to slightly inferior accuracy compared to a depth of 12. It is observed that when increasing the layer depth from front to back, the performance improves consistently until saturation. 

\textbf{Prompt Length.} As illustrated in Figure \ref{fig:img2}, we present the influence of prompt length within the \ChordPrompt framework. An interesting pattern emerges where the performance on the test set relative to the original classes declines as the prompt length increases. This trend indicates that longer prompts may lead to overfitting, which harms the model's ability to generalize to unseen test samples. Therefore, we use relatively shorter prompts that may provide an ideal balance between learnability and generalizability. These findings highlight the importance of carefully tuning layer depth and length to balance model adaptability and generalization capability in continual learning scenarios.

\textbf{Effectiveness of Domain-Adaptive Text Prompt.}
The effectiveness of our domain-adaptive text prompt is demonstrated in Table \ref{table:table4}, where ChordPrompt significantly outperforms other prompt-based methods such as L2P (-13.2\% in Transfer accuracy), DualPrompt (-13.0\%), and S-Prompts (-13.2\%). These competing prompt methods lack domain-adaptive capabilities, resulting in their inability to select appropriate prompts for different domains, which explains their substantial performance degradation in multi-domain scenarios. MaPLe, despite using multi-modal prompts, still falls short in the Last accuracy metric (83.9\% vs. our 86.0\%) due to its unidirectional information flow and absence of domain-adaptive mechanisms for prompt selection across varied domains.

\begin{table}[!ht]

\centering
\fontsize{10}{10}\selectfont
\setlength{\tabcolsep}{8pt}  
\renewcommand{\arraystretch}{1.5}  
\begin{tabular}{lcccccc}
\hline
Method & Transfer & $\Delta$ & Avg. & $\Delta$ & Last & $\Delta$ \\
\hline
Zero-shot & 65.4 & 0.0 & 65.3 & 0.0 & 65.3 & 0.0 \\
CL. & 46.6 & -18.8 & 56.2 & -9.1 & 67.4 & 2.1 \\
LwF & 53.2 & -12.2 & 62.2 & -3.1 & 71.9 & 6.6 \\
iCaRL & 50.9 & -14.5 & 56.9 & -8.4 & 71.6 & 6.3 \\
LwF-VR & 53.1 & -12.3 & 60.6 & -4.7 & 68.3 & 3.0 \\
WiSE-FT & 51.0 & -14.4 & 61.5 & -3.8 & 72.2 & 6.9 \\
ZSCL & 64.2 & -1.2 & 74.5 & 9.2 & 83.4 & 18.1 \\
\rowcolor{gray!25}
\textbf{\ChordPrompt} & \textbf{65.4} & \textbf{0.0} & \textbf{75.6} & \textbf{10.3} & \textbf{85.1} & \textbf{19.8} \\
\hline
\end{tabular}
\caption{Compare methods on MTIL in Order II.}
\label{table:table5}
\end{table}
\textbf{Parameter Analysis.}
Parameter Analysis in Table \ref{tab:mtil} showcases the computational complexity comparison among different methods. \ChordPrompt introduces 9.48M trainable parameters, \ChordPrompt's trainable parameter count is substantially lower than methods like iCaRL, LWF, and ZSCL, which require fine-tuning the entire CLIP model (211M parameters). 
\ChordPrompt substantially improves in continual learning scenarios while only requiring updates to a small portion of the overall model parameters, making it more efficient than full model fine-tuning approaches. 
This lightweight design ensures compatibility with resource-constrained environments while maintaining superior performance in continual learning scenarios.

\textbf{Few-shot Multi-Domain Task Incremental Learning.}
We further assess our method in a few-shot multi-domain task incremental learning scenario, where the CLIP model is restricted to only a handful of samples per task. Under the 5-shot setting, Table \ref{tab:few_shot_mtil_comparison} presents the results based on the same evaluation metrics as Table \ref{tab:mtil}. Our approach consistently achieves better performance than leading existing methods across most datasets. These findings indicate that the proposed methods are highly effective at mitigating the forgetting problem in continual learning. Moreover, the domain-adaptive prompt in our framework is able to efficiently distinguish between data distributions of different tasks, demonstrating robust distribution discrimination capabilities even with limited training examples.

\textbf{ Other Prompt Methods and Ablation Study.}
Our experimental results demonstrate that the cross-modal prompt approach consistently outperforms other continual learning methods across various tasks. 

\begin{table}[htb]

\centering
\fontsize{10}{10}\selectfont
\setlength{\tabcolsep}{8pt}  
\renewcommand{\arraystretch}{1.5}  
\begin{tabular}{lcccccc}
\hline
Method & Transfer & $\Delta$ & Avg. & $\Delta$ & Last & $\Delta$ \\
\hline
Zero-shot & 69.5 & 0.0 & 65.3 & 0.0 & 65.3 & 0.0 \\
L2P & 53.2 & -16.3 & 67.9 & 2.6 & 82.0 & 16.7 \\
Dualp. & 52.4 & -17.1 & 68.0 & 2.7 & 82.3 & 17.0 \\
S-Prompts & 52.2 & -17.3 & 68.3 & 3.0 & 82.4 & 17.1 \\
\hline

\rowcolor{gray!25}
\textbf{\ChordPrompt} & \textbf{69.4} & \textbf{-0.1} & \textbf{77.2} & \textbf{11.9} & \textbf{86.0} & \textbf{20.7} \\
\hline
\end{tabular}
\caption{Ablation Study on MTIL in Order I.}
\label{table:table4}
\end{table}

In Table \ref{table:table4}, L2P \cite{wang2022learning}, DualPrompt \cite{wang2022dualprompt}, and S-Prompts \cite{wang2022s} all use single-modal prompts without domain-adaption. This neglects the benefits of cross-modal synergy and domain feature. As a result, these methods perform poorly in multi-domain task incremental learning (MTIL) scenarios, as they were primarily designed for class-incremental learning and fail to adapt effectively across different domains.

\ChordPrompt's superior performance stems from its unique cross-modal information sharing mechanism. Unlike traditional methods that allow multi-modal interaction at the final stage, \ChordPrompt shares prompt information across all CLIP model layers. This cross-layer sharing of prompts enables a more comprehensive exchange of information between the visual and textual modalities, allowing the model to capture fine-grained correspondences and interactions at various levels of abstraction. By facilitating this deep integration of cross-modal information, \ChordPrompt can better align the visual and textual representations, leading to more robust and adaptable continual learning. The performance gain is particularly notable in complex, multi-domain tasks like the MTIL benchmark, demonstrating \ChordPrompt's effectiveness in integrating diverse knowledge types. This underscores its potential for real-world applications with blurred task boundaries and crucial multi-modal processing.

\section{Conclusion}
\ChordPrompt enhances the deployment of vision-language models by addressing continual learning challenges and eliminating the need for costly retraining. Its strong performance in zero-shot transfer and downstream tasks highlights its practicality and versatility. Our experiments validate the effectiveness of domain-adaptive text and cross-modal visual prompts in preserving task-specific and general knowledge.

While achieving significant improvements in classification tasks, future work will explore extending ChordPrompt to generative vision-language models, such as visual question answering and image captioning, to handle more open-ended and complex multi-modal interactions.



\bibliographystyle{splncs04}


\end{document}